\documentclass{article}
\usepackage{spconf,amsmath,graphicx}
\usepackage[named]{algo}
\usepackage{amsfonts}
\usepackage{amssymb}
\usepackage{amsmath}
\usepackage{amsthm}
\theoremstyle{remark}
\newtheorem{remark}{Remark}
\usepackage{hyperref}

\title{A short tutorial on the Weisfeiler-Lehman test and its variants}
\name{Ningyuan (Teresa) Huang, Soledad Villar\thanks{SV is supported by NSF DMS 2044349, EOARD FA9550-18-1-7007, and the NSF-Simons Research Collaboration on the Mathematical and Scientific Foundations of Deep Learning (MoDL) (NSF DMS 2031985). The authors thank Zhengdao Chen for motivating this note and pointing out relevant references; Haoteng Yin, Rucha Joshi for pointing out errors in a previous version of this manuscript; and David W. Hogg for writing advice.}}
\address{
Applied Mathematics and Statistics\\
Mathematical Institute for Data Science\\
Johns Hopkins University}
\begin{document}
%
\maketitle
\begin{abstract}
Graph neural networks are designed to learn functions on graphs. Typically, the relevant target functions are invariant with respect to actions by permutations. Therefore the design of some graph neural network architectures has been inspired by graph-isomorphism algorithms. 

The classical Weisfeiler-Lehman algorithm (WL)---a graph-isomorphism test based on color refinement---became relevant to the study of graph neural networks. The WL test can be generalized to a hierarchy of higher-order tests, known as $k$-WL. This hierarchy has been used to characterize the expressive power of graph neural networks, and to inspire the design of graph neural network architectures. 

A few variants of the WL hierarchy appear in the literature. The goal of this short note is pedagogical and practical: We explain the differences between the WL and folklore-WL formulations, with pointers to existing discussions in the literature. We illuminate the differences between the formulations by visualizing an example.

\end{abstract}
\begin{keywords}
Graph neural networks, Weisfeiler Lehman
\end{keywords}
\section{Introduction}

In the past few years, deep learning has completely revolutionized entire fields: Convolutional neural networks have changed the landscape of computer vision, and recurrent neural networks significantly improved the state of the art in natural language processing \cite{lecun2015deep}. Deep learning is now being applied, with different degrees of success, to more general problems and datasets, arising from scientific and industrial applications. There is a natural flow in the field towards the study of geometric deep learning beyond Euclidean data \cite{bronstein2017geometric}, where the network architecture encodes relevant theoretical properties of the problems they are trying to solve (symmetries, invariances, conservation laws). This is best exemplified by data structures like manifolds and graphs. 

Graphs are one of the most common abstractions to represent data, and unsurprisingly, the study of graph neural networks (GNNs) is a rapidly growing field. There are many types GNN architectures, based on message passing \cite{gilmer2017neural, hamilton2017representation}, convolutional filters \cite{bruna2013spectral, gama2020graphs}, generalization of spectral methods \cite{chen2019cdsbm}, invariant linear functions \cite{maron2018invariant, maron2019provably}, to name a few.   

Graph neural networks are typically formulated as functions that take a graph as input and output a representation of the graph. The representation usually takes the form of an embedding of the graph nodes in Euclidean space. One fundamental property of most graph neural networks is the invariance (or equivariance) with respect to permutations of the input.  The philosophy is that the learned representation of the graph should be consistent with any relabeling of the nodes. This sometimes restricts the class of functions that graph neural networks can express, and constrains the architectural design of the neural network. In a nutshell, there seems to be a trade-off between invariance and expressibility. Being able to approximate invariant functions is closely related to being able to decide whether any pair of graphs are isomorphic, which is not an easy problem \cite{chen2019equivalence}. 

Graph isomorphism is a long-standing problem in theoretical computer science. It was suspected to be NP-hard until quite recently, when L\'azl\'o Babai produced a quasi-polynomial time algorithm to decide whether two graphs are isomorphic \cite{babai2016graph}. There are other, less sophisticated, tests for graph isomorphism that don't fully characterize graphs modulo isomorphisms, but can distinguish large sets of non-isomorphic graphs.  

The Weisfeiler-Lehman (WL) algorithm is a classical isomorphism test based on color refinement \cite{weisfeiler1968reduction}. Each node keeps a state (or color) that gets refined in each iteration by aggregating information from their neighbor's states. The refinement stabilizes after a few iterations and it outputs a representation of the graph. Two graphs with different representations are not isomorphic. The test can uniquely identify a large set of graphs up to isomorphism \cite{babai1979canonical}, but there are simple examples where the test tragically fails---for instance, two regular graphs with the same number of nodes and same degrees cannot be distinguished by the test, even if one is connected and the other one is not.

A natural extension of the test provides a hierarchy of algorithms. Instead of keeping the state of one node, they keep the state of $k$-tuples of nodes. These algorithms are called $k$-dimensional Weisfeiler-Lehman tests or $k$-WL. There are two main versions of the algorithm with slightly different update rules, the one studied in \cite{cai1992optimal} has recently been named $k$-folklore-WL ($k$-FWL) \cite{morris2019higher}, while the one studied in \cite{grohe2015pebble, grohe2017descriptive} is known as $k$-WL.  

The power of $k$-WL and $k$-FWL to distinguish non-isomorphic graphs is well understood. Very beautiful mathematical work has characterized their discriminative power in terms of the satisfiability of quantified logical formulas on finite variables and pebble games \cite{cai1992optimal, grohe2015pebble}, and in terms of the Sherali-Adams linear programming hierarchy \cite{grohe2015pebble, atserias2013sherali}. We refer the reader to the very comprehensive book by Martin Grohe \cite{grohe2017descriptive}. 

The WL hierarchy of graph isomorphism tests has shown to be a great inspiration to define functions on graphs, in particular graph kernels \cite{shervashidze2011weisfeiler, togninalli2019wasserstein}, and graph neural network architectures \cite{morris2019higher, morris2019towards, maron2019provably}; and it has been proven to be a powerful tool to theoretically analyze the expressive power of graph neural networks \cite{xu2018powerful, morris2019higher, chen2019equivalence}. 

In this short note, we explicitly state the differences between the test formulations, and we point out the literature of graph neural network inspired by the WL hierarchy, complementing the very nice survey by Sato \cite{sato2020survey}.

\label{sec:intro}

\section{Setting and notation}
We consider undirected graphs $G=(V,E,X_V)$, where $V=\{1,\ldots n\}:=[n]$ is the set of vertices; $E\subseteq V\times V$ is the set of vertices satisfying $(u,v)\in E$ if and only if $(v,u)\in E$; and $X_V$ is the set of node features: For all $v \in V$ $X_v\in \mathbb R^{d}$. The neighbors of a vertex $v$ is $\mathcal N_G(v)=\{w: (v,w)\in E\}$. Graphs without node features can be represented either by taking $X_v=1$ for all $v\in V$, or assigning a unique identifier for each node. In Section \ref{sec:example}, we focus on the former setting where nodes are \textit{anonymous}, followed from the original formulation of WL algorithm. However, we do point out that the latter setting is shown to be more powerful by allowing \textit{non-anonymous} node features \cite{Loukas2020What}.

We say $G=(V,E,X_V)$ and $G'=(V',E',X'_V)$ are isomorphic if there exists a relabeling of the nodes of $G$ that produce the graph $G'$. In other words, they are isomorphic if there exists a permutation $\Pi \in S_n$ so that $\Pi\,V=V'$, $\Pi\,E=E'$ where $\Pi\,(u,v):=(\Pi\,u, \Pi\,v)$ and $\Pi \,X_V = X'_V$ where $\Pi X_v = X_{\Pi v}$.  

The Weisfeiler-Lehman test keeps a state (or color) for every node (or tuples of nodes denoted by $\vec v =(v_1,\ldots, v_k) \in V^k$ in its $k$-dimensional versions). It refines the node states by aggregating the state information from their neighbors. In order to compute the update, WL uses an injective hash function defined in different objects modulo equivalence classes. In particular, for all $v,w\in V$ $\text{hash}(X_v)=\text{hash}(X_w)$ iff $X_v=X_w$. For $\vec v =(v_1,\ldots v_k)$, $\vec v' =(v'_1,\ldots v'_k)$ we define the hash function such that  
$\text{hash}(G[\vec v])=\text{hash}(G[\vec {v'}])$ iff (1) $X_{v_i}=X_{v'_i} \forall i \in [k]$;  and (2) $(v_i,v_j)\in E$ iff $(v'_i,v'_j)\in E', \forall i,j \in [k]$.  

We use the notation $\{\!\!\{\cdot\}\!\!\}$ to denote a multiset. Two multisets are equal (and have the same hash value) if they have the same elements with equal multiplicities. In practice,  this is implemented by sorting the elements and applying the hash function on the sorted values. Two tuples have the same hash value if and only if the hash of their respective (ordered) entries coincide. Finally, two states produce the same refinement $(c_v^\ell)_{v\in V}=(c_v^{\ell'})_{v\in V}$ if the level sets coincide.





\section{Weisfeiler Lehman variants}
\label{sec:WL}

\subsection{1-WL (color refinement)}
The classical WL (or 1-WL) test \cite{weisfeiler1968reduction}, keeps a state for each node that refines by aggregating their neighbors state. It outputs an embedding of the graph that corresponds to the state of every node. We say that the WL succeeds at distinguishing a pair of non-isomorphic graphs $G,G'$ if $\text{WL}(G)\neq \text{WL}(G')$. 

\begin{algorithm}{1-WL (color refinement)}{
\label{algo:ifforwhile}
\qinput{$G=(V,E,X_V)$}}
$c_v^{0} \qlet$ hash$(X_v)$ for all $v\in V$ \\
\qrepeat \\
$c_v^\ell \qlet$ hash$(c_v^{\ell-1}, \{\!\!\{c_w^{\ell-1} : w \in \mathcal N_G(v)\}\!\!\}) \; \forall v\in V$
\quntil $(c_v^\ell)_{v\in V}= (c_v^{\ell-1})_{v \in V}$\\
\qreturn $\{\!\!\{c_v^\ell: v \in V\}\!\!\}$
\end{algorithm}

The WL algorithm successfully distinguishes most pairs of graphs \cite{babai1979canonical}, but it fails to distinguish some basic examples, such as all regular graphs on $n$ nodes and degree $d$. In the context of GNNs, \cite{morris2019higher} and \cite{xu2018powerful} show that under the \textit{anonymous} setting, if $f_\theta$ is a function implemented by a message passing neural network (MPNN), then $f_\theta(G)=f_\theta(G')$ for all $G,G'$ such that $\text{WL}(G)=\text{WL}(G')$. In particular, MPNNs cannot express some trivial functions, such as the number of connected components of a graph, given that the graph does not have node features and $X_v$ is taken as the same for all nodes.

\subsection{$k$-WL}
\label{sec:kwl}
The $k$-dimensional Weisfeiler Lehman test extends the test to coloring $k$-tuples of nodes. It is defined as:
\begin{algorithm}{$k$-WL ($k\geq 2$)}{
\label{algo:ifforwhile}
\qinput{$G=(V,E,X_V)$}}
$c_{\vec v}^{0} \qlet$ hash$(G[\vec v])$ for all $\vec v\in V^k$ \\
\qrepeat \\
$c_{\vec v, i}^\ell \qlet  \{\!\!\{c_w^{\ell-1} : w \in  \mathcal{N}_{i}(\vec v) \}\!\!\}\;  \forall v \in V^k, i\in[k]$\\
$c_{\vec v}^\ell \qlet$ hash$(c_{\vec v}^{\ell-1}, c_{\vec v, 1}^{\ell}, \ldots c_{\vec v, k}^{\ell}) \; \forall \vec{v} \in V^k$
\quntil $(c_{\vec v}^\ell)_{\vec v\in V^k}== (c_{\vec v}^{\ell-1})_{\vec v\in V^k}$ \\
\qreturn $\{\!\!\{c_{\vec v}^\ell: \vec v \in V^k\}\!\!\}$
\end{algorithm}

The node neighborhood $\mathcal{N}_{i}(\vec v)$ is the set of $k$-tuples that differ with $\vec v$ only in the position $i$. For $\vec v =(v_1, \ldots, v_k)$ we have
$$
\mathcal{N}_{i}(\vec v) =\left\{\left(v_{1}, \ldots, v_{i-1}, w, v_{i+1}, \ldots, v_{k}\right) : w \in V\right\}.
$$

Inspired by k-WL, \cite{morris2019higher} propose a GNN architecture based on a version of $k$-WL on sets, which is strictly more expressive than MPNN. 

\subsection{$k$-FWL}
The version of the $k$-Weisfeiler Lehman test studied by Cai, Furer, and Immmerman \cite{cai1992optimal} considers a slightly different definition for the updates than the definition in Section~\ref{sec:kwl}. It has recently been renamed as folklore-WL (FWL) by \cite{morris2019higher}. It is computationally more efficient than $k$-WL and it has been used in \cite{maron2019provably, chen2019equivalence, damke2020novel} to design GNN architectures. 

\begin{algorithm}{$k$-FWL ($k\geq 2$)}{
\label{algo:ifforwhile}
\qinput{$G=(V,E,X_V)$}}
$c_{\vec v}^{0} \qlet$ hash$(G[\vec v])$ for all $\vec v\in V^k$ \\
\qrepeat \\
$c_{\vec v, w}^\ell \qlet  (c_{\vec v[1]\leftarrow w}^{\ell-1} \ldots, c_{\vec v[k]\leftarrow w}^{\ell-1}); \forall v \in V^k, w\in V$\\
$c_{\vec v}^\ell \qlet$ hash$(c_{\vec v}^{\ell-1}, \{\!\!\{ c_{\vec v, w}^{\ell} : w\in V\}\!\!\}) \; \forall \vec{v} \in V^k$
\quntil $(c_{\vec v}^\ell)_{\vec v \in V^k}= (c_{\vec v}^{\ell-1})_{\vec v \in V^k}$\\
\qreturn $\{\!\!\{c_{\vec v}^\ell: \vec v \in V^k \}\!\!\}$
\end{algorithm}

Note that $c_{\vec v[i]\leftarrow w}$ is a tuple that differs from $\vec v$ in the position $i$, where $v_i$ is exchanged by $w$. In particular, if $\vec v =(v_1, \ldots, v_k)$ then
$$c_{\vec v[i]\leftarrow w}= (v_1, \ldots, v_{i-1}, w, v_{i+1}, \ldots v_k).$$



Thus, the node neighborhood $\mathcal{N}_i^F(\vec v)$ in $k$-FWL is
$$
\mathcal{N}_{i}^{F}(\vec v) =\left(\left(i, v_{2}, \ldots \right),\left(v_{1}, i, \ldots \right), \ldots,\left(\ldots, v_{k-1}, i\right)\right) 
$$

We can see that in $k$-WL, $\mathcal{N}_i (\vec v)$ is a set of $n$ elements where each element is $k$-dimensional; in $k$-FWL, $\mathcal{N}^F_i (\vec v)$ is a set of $k$ elements where each element is $n$-dimensional. The definition of neighborhood underpins the differences in $k$-WL and $k$-FWL. More explanations and illustrations are given in Section \ref{sec:example}.




\subsection{Comparisons between the WL variants} \label{sec:comparison}

\begin{remark}
Note that 1-WL is not the same as $k$-WL with $k=1$. In fact, 1-WL$\equiv$2-WL. 
\end{remark}
\begin{proof}
Consider $k$-WL defined in Section \ref{sec:kwl} with $k=1$: then $c^1_{\vec v,1}$ would be the same for all $\vec v$ (the multiset of $c^0_{\vec v}$ for $\vec v\in V^1$), so the algorithm stabilizes in one step, coinciding with the initialization. This is because in $k$-WL the neighboring tuples do not have information about the edges of the graph. The edges are only considered in the initialization hash$(G[\vec v])$.


Let $G=(V,E,X_V)$, $\vec v =(v_i, v_j)\in V^2$ and $\vec v'=(v_k, v_l)\in V^2$. After one step of 2-WL we have  $c^1_{\vec v}= c^1_{\vec v'}$ if and only if (1) $c^0_{\vec v}= c^0_{\vec v'}$, (2) $\{\!\!\{ c^0_{w, v_j}: w \in V \}\!\!\}= \{\!\!\{ c^0_{w, v_l}: w \in V \}\!\!\}$, and (3) $\{\!\!\{ c^0_{v_i, w}: w \in V \}\!\!\}= \{\!\!\{ c^0_{v_k, w}: w \in V \}\!\!\}$. Note that (2) holds if and only if $ \{\!\!\{ \text{hash}(w): w\in \mathcal N(v_j)  \}\!\!\}= \{\!\!\{ \text{hash}(w): w\in \mathcal N(v_l)  \}\!\!\}$ and similarly for (3).

This argument shows that in each iteration, color refinement in 2-WL is equivalent to implementing 1-WL in each coordinate. This argument inductively shows that $\text{2-WL}(v_i,v_j)=(\text{1-WL}(v_i),\text{1-WL}(v_j))$ and therefore it has the same distinguishing power as 1-WL.

\end{proof}

\begin{remark}
The discriminating power of $k$-WL is equivalent to the one of $(k-1)$-FWL for $k\geq 3$. To the best of our knowledge, there is no explicit proof of the equivalence only relying on the definitions of $k$-WL and $(k-1)$-WL. However, in Section 5 of \cite{cai1992optimal} the authors prove that $(k-1)$-FWL is equivalent to $C^k$, the set of quantified first order formulas of $G=(V,E)$ in $k$ variables\footnote{One example of such formulas is $\forall x_1 \exists!d x_2(E(x_1,x_2))$. This means that for all $x_1$ node in $G$, there exists exactly $d$ nodes $x_2$ ($!$ means exactly) such that there is an edge between $x_1$ and $x_2$ (i.e. the graph $G$ has degree $d$).}.  This proof is reformulated in Theorem 3.5.7 of \cite{grohe2017descriptive} for $k$-WL, showing that $k$-WL is equivalent to $C^k$.
\end{remark}

\begin{remark}
For some applications it makes sense to consider weighted graphs, or more generally, graphs with edge features $G=(V,E,X_V, X_E)$ where $X_e\in \mathbb R^{m}$ for $e\in E$. Although the classical WL test is not equipped to deal with edge features a priori, the higher dimensional versions of the test can be easily extended to work with edge features by extending the definition of hash$(G[\vec v])$ to consider edge features.
\end{remark}

\subsection{Example: $2$-WL and $2$-FWL on regular graphs}\label{sec:example}

\begin{figure}[ht]
\begin{center}
\includegraphics[width=0.35\textwidth]{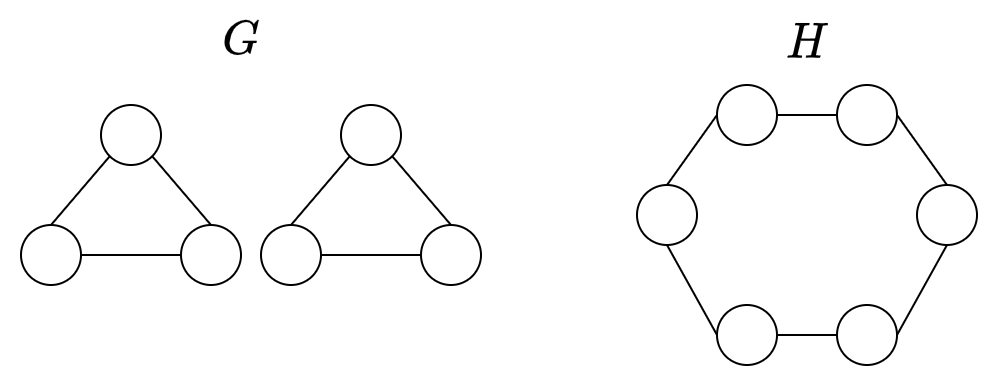}
\end{center}
\vspace{-.8cm}
\caption{Two graphs can not be distinguished by WL and $2$-WL, but can be distinguished by $2$-FWL.} 
\label{example}
\end{figure}

Below we show a canonical example of two regular non-isomorphic graphs (Figure \ref{example}), where the classical WL test and $2$-WL test both fail to distinguish, but $2$-FWL succeeds.  

\begin{figure}[t]
\begin{center}
\includegraphics[width=0.49\textwidth]{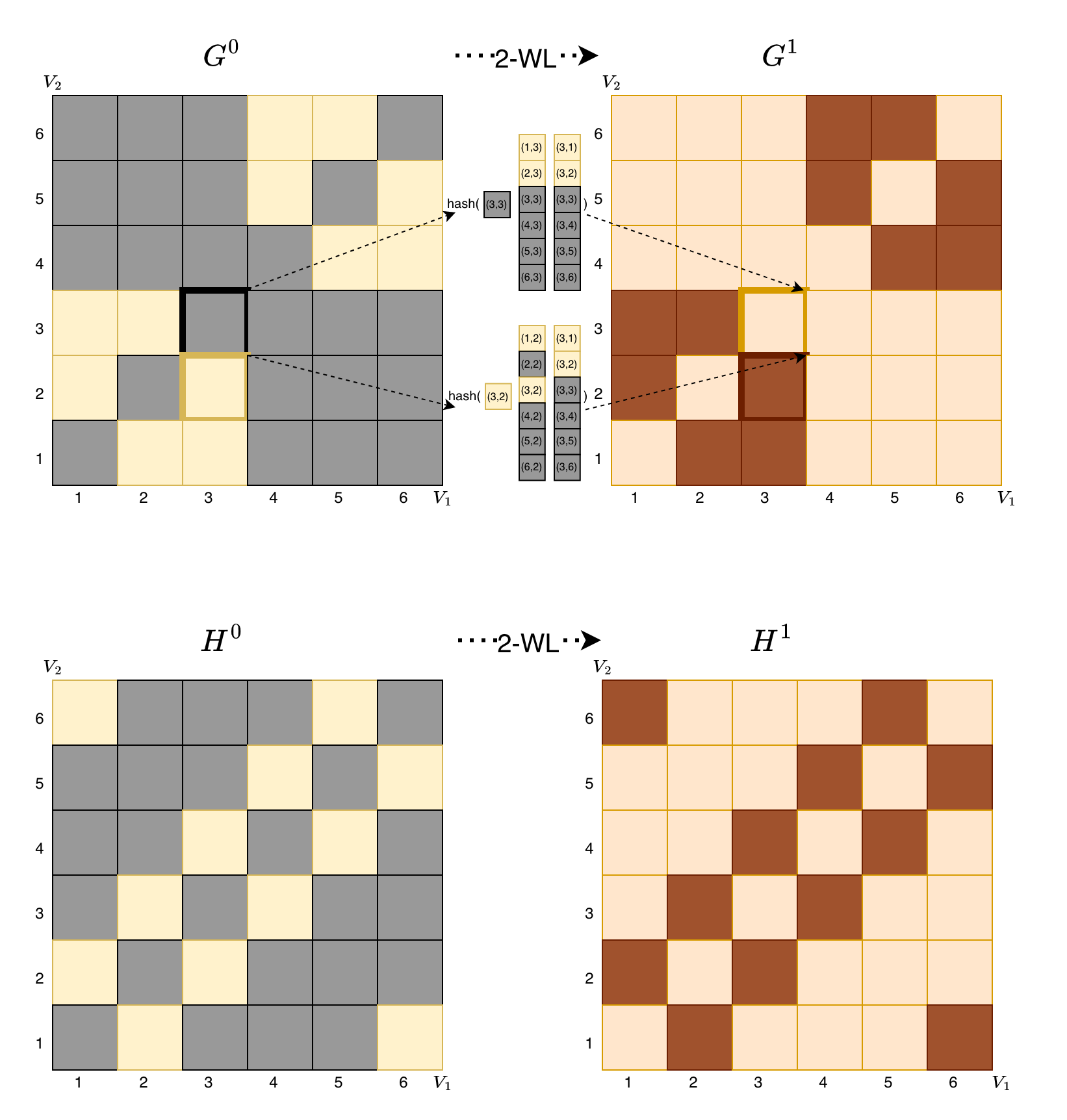}
\end{center}
\vspace{-.7cm}
\caption{$2$-WL color assignment at initialization $G^0, H^0$ and refinement at first round $G^1, H^1$. The color refinement stabilizes in one step, failing to distinguish $G$ and $H$} 
\label{2-WL}
\end{figure}

We first go over the steps for $2$-WL, illustrated in Figure \ref{2-WL}. For simplicity we assume that all node features are the same. At the initialization, there are only two isomorphic types: (1) $(v_i, v_j) \in E$ (i.e, connected); (2) $(v_i, v_j) \notin E$ (i.e., not connected). We color $\vec v$ as yellow for type (1) and grey for type (2). As shown in Figure \ref{2-WL} left panel, $G^0$ and $H^0$ (the coloring at initialization) has the same elements with equal multiplicities (24 greys; 12 yellows), and thus they produce the same hash value. Now, we can view $V^2$ as a $6 \times 6$ matrix, and the neighbors of a tuple $(v_i, v_j)$ is given by the $j$-th row and the $i$-th column. Examples of neighbors for $(3,3), (3,2)$ are shown in middle inset of Figure \ref{2-WL}. Observe that for any node in $G^0$, $c_{\vec v, 1}^{1} = c_{\vec v, 2}^{1} = \{\!\!\{ 4 \text{ greys}, 2 \text{ yellows} \}\!\!\}$. Thus all nodes have the same neighborhood but may differ in the initial color. Let hash$(\text{grey}, c_{\vec v, 1}^{1}, c_{\vec v, 2}^{1})$ be orange, hash$(\text{yellow}, c_{\vec v, 1}^{1}, c_{\vec v, 2}^{1})$ be brown, and we obtain $G^1, H^1$ as shown on the right panel of Figure \ref{2-WL}. Notice that the color patterns (i.e., the multiset of $G, H$) do not change after the first iteration, and thus the $2$-WL test terminates, which fails to distinguish $G$ and $H$.

\begin{figure}[t]
\begin{center}
\includegraphics[clip, trim=0cm 0cm 0cm 0.5cm, width=0.4\textwidth]{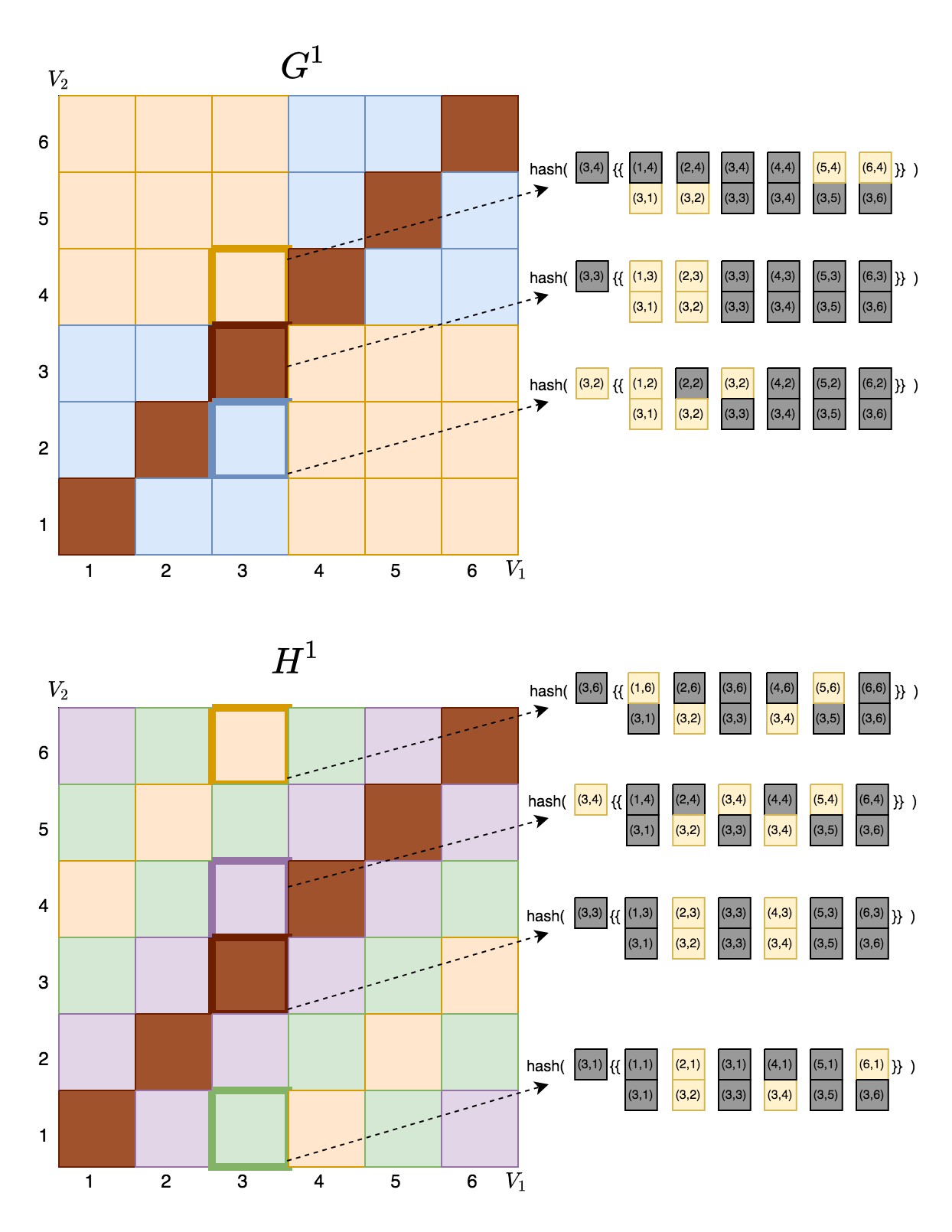}
\end{center}
\vspace{-.8cm}
\caption{$2$-FWL color refinement at first round $G^1, H^1$ outputs different color patterns for $G$ and $H$. Thus it  distinguishes $G$ and $H$, and terminates in one iteration.} 
\label{2-FWL}
\end{figure}

Figure \ref{2-FWL} illustrates the color refinement of $2$-FWL. Note that $2$-FWL has the same color assignment as $2$-WL at the initialization. However, $2$-FWL defines the tuple neighborhood as $n$ elements of length $2$, unlike $2$ vectors of length $n$ in $2$-WL. Examples of neighbors defined by $2$-FWL are shown at the right panel of Figure \ref{2-FWL}. Now, there are 3 isomorphic types in $G^0$, which are hashed as brown, blue, and orange. Intuitively, they represent $0$-hop, $1$-hop, and disjoint neighborhood in $G$, respectively. In contrast, $H$ has 4 isomorphic types hashed as brown, purple, green, and orange, which characterizes the $0,1,2,3$-hop neighborhood in $H$. Hence, in the first iteration, $2$-FWL outputs two different color patterns for $G$ and $H$. One can check the refinement stabilizes in one step, correctly concluding that $G$ and $H$ are non-isomorphic.

\section{Conclusion}
\label{sec:beyond}



The Weisfeler-Lehman test and its $k$-dimensional generalizations are powerful tools to study the expressive power of invariant functions on graphs. Mathematically, this test is very well understood thanks to the work by Cai, Furer, and Immerman in the 90's and the work by Grohe and collaborators in the past decade. 

In the context of graph neural networks, the Weisfeiler-Lehman test has been used to analyze their theoretical properties. It has also inspired the design of expressive and invariant GNN architectures. 

In this work, we explain the context of $k$-WL and $k$-FWL and provide a simple tutorial that illustrates the differences between them. We refer the reader to \cite{sato2020survey} for a comprehensive survey.



\label{sec:refs}

\bibliography{ref}

\begin{thebibliography}{10}

\bibitem{lecun2015deep}
Yann LeCun, Yoshua Bengio, and Geoffrey Hinton,
\newblock ``Deep learning,''
\newblock {\em nature}, vol. 521, no. 7553, pp. 436--444, 2015.

\bibitem{bronstein2017geometric}
M.~M. {Bronstein}, J.~{Bruna}, Y.~{LeCun}, A.~{Szlam}, and P.~{Vandergheynst},
\newblock ``Geometric deep learning: Going beyond euclidean data,''
\newblock {\em IEEE Signal Processing Magazine}, vol. 34, no. 4, pp. 18--42,
  2017.

\bibitem{gilmer2017neural}
Justin Gilmer, Samuel~S Schoenholz, Patrick~F Riley, Oriol Vinyals, and
  George~E Dahl,
\newblock ``Neural message passing for quantum chemistry,''
\newblock in {\em Proceedings of the 34th International Conference on Machine
  Learning-Volume 70}. JMLR. org, 2017, pp. 1263--1272.

\bibitem{hamilton2017representation}
William~L Hamilton, Rex Ying, and Jure Leskovec,
\newblock ``Representation learning on graphs: Methods and applications,''
\newblock {\em arXiv preprint arXiv:1709.05584}, 2017.

\bibitem{bruna2013spectral}
Joan Bruna, Wojciech Zaremba, Arthur Szlam, and Yann LeCun,
\newblock ``Spectral networks and locally connected networks on graphs,''
\newblock {\em arXiv preprint arXiv:1312.6203}, 2013.

\bibitem{gama2020graphs}
Fernando Gama, Elvin Isufi, Geert Leus, and Alejandro Ribeiro,
\newblock ``Graphs, convolutions, and neural networks,''
\newblock {\em IEEE SIGNAL PROCESSING MAGAZINE}, vol. 1053, no. 5888/20, 2020.

\bibitem{chen2019cdsbm}
Zhengdao Chen, Lisha Li, and Joan Bruna,
\newblock ``Supervised community detection with line graph neural networks,''
\newblock {\em Internation Conference on Learning Representations}, 2019.

\bibitem{maron2018invariant}
Haggai Maron, Heli Ben-Hamu, Nadav Shamir, and Yaron Lipman,
\newblock ``Invariant and equivariant graph networks,''
\newblock 2018.

\bibitem{maron2019provably}
Haggai Maron, Heli Ben-Hamu, Hadar Serviansky, and Yaron Lipman,
\newblock ``Provably powerful graph networks,''
\newblock in {\em Advances in Neural Information Processing Systems}, 6 2019,
  pp. 2153--2164.

\bibitem{chen2019equivalence}
Zhengdao Chen, Soledad Villar, Lei Chen, and Joan Bruna,
\newblock ``On the equivalence between graph isomorphism testing and function
  approximation with gnns,''
\newblock in {\em Advances in Neural Information Processing Systems}, 2019, pp.
  15868--15876.

\bibitem{babai2016graph}
L{\'a}szl{\'o} Babai,
\newblock ``Graph isomorphism in quasipolynomial time,''
\newblock in {\em Proceedings of the forty-eighth annual ACM symposium on
  Theory of Computing}. ACM, 2016, pp. 684--697.

\bibitem{weisfeiler1968reduction}
B~Weisfeiler and A~Leman,
\newblock ``The reduction of a graph to canonical form and the algebra which
  appears therein,''
\newblock {\em Nauchno-Technicheskaya Informatsia}, vol. 2(9):12-16, 1968.

\bibitem{babai1979canonical}
L{\'a}szl{\'o} Babai and Ludik Kucera,
\newblock ``Canonical labelling of graphs in linear average time,''
\newblock in {\em 20th Annual Symposium on Foundations of Computer Science
  (sfcs 1979)}. IEEE, 1979, pp. 39--46.

\bibitem{cai1992optimal}
Jin-Yi Cai, Martin F{\"u}rer, and Neil Immerman,
\newblock ``An optimal lower bound on the number of variables for graph
  identification,''
\newblock {\em Combinatorica}, vol. 12, no. 4, pp. 389--410, 1992.

\bibitem{morris2019higher}
Christopher Morris, Martin Ritzert, Matthias Fey, William~L Hamilton, Jan~Eric
  Lenssen, Gaurav Rattan, and Martin Grohe,
\newblock ``Weisfeiler and leman go neural: Higher-order graph neural
  networks,''
\newblock {\em Association for the Advancement of Artificial Intelligence},
  2019.

\bibitem{grohe2015pebble}
Martin Grohe and Martin Otto,
\newblock ``Pebble games and linear equations,''
\newblock {\em The Journal of Symbolic Logic}, pp. 797--844, 2015.

\bibitem{grohe2017descriptive}
Martin Grohe,
\newblock {\em Descriptive complexity, canonisation, and definable graph
  structure theory}, vol.~47,
\newblock Cambridge University Press, 2017.

\bibitem{atserias2013sherali}
Albert Atserias and Elitza Maneva,
\newblock ``Sherali--adams relaxations and indistinguishability in counting
  logics,''
\newblock {\em SIAM Journal on Computing}, vol. 42, no. 1, pp. 112--137, 2013.

\bibitem{shervashidze2011weisfeiler}
Nino Shervashidze, Pascal Schweitzer, Erik~Jan Van~Leeuwen, Kurt Mehlhorn, and
  Karsten~M Borgwardt,
\newblock ``Weisfeiler-lehman graph kernels.,''
\newblock {\em Journal of Machine Learning Research}, vol. 12, no. 9, 2011.

\bibitem{togninalli2019wasserstein}
Matteo Togninalli, Elisabetta Ghisu, Felipe Llinares-L{\'o}pez, Bastian Rieck,
  and Karsten Borgwardt,
\newblock ``Wasserstein weisfeiler-lehman graph kernels,''
\newblock in {\em Advances in Neural Information Processing Systems}, 2019, pp.
  6439--6449.

\bibitem{morris2019towards}
Christopher Morris and Petra Mutzel,
\newblock ``Towards a practical $ k $-dimensional weisfeiler-leman algorithm,''
\newblock {\em arXiv preprint arXiv:1904.01543}, 2019.

\bibitem{xu2018powerful}
Keyulu Xu, Weihua Hu, Jure Leskovec, and Stefanie Jegelka,
\newblock ``How powerful are graph neural networks?,''
\newblock {\em arXiv preprint arXiv:1810.00826}, 2018.

\bibitem{sato2020survey}
Ryoma Sato,
\newblock ``A survey on the expressive power of graph neural networks,''
\newblock {\em arXiv preprint arXiv:2003.04078}, 2020.

\bibitem{Loukas2020What}
Andreas Loukas,
\newblock ``What graph neural networks cannot learn: depth vs width,''
\newblock in {\em International Conference on Learning Representations}, 2020.

\bibitem{damke2020novel}
Clemens Damke, Vitalik Melnikov, and Eyke H{\"u}llermeier,
\newblock ``A novel higher-order weisfeiler-lehman graph convolution,''
\newblock in {\em Asian Conference on Machine Learning}. PMLR, 2020, pp.
  49--64.

\end{thebibliography}
\bibliographystyle{IEEEbib}

\end{document}